\UseRawInputEncoding
\documentclass[runningheads]{llncs}
\usepackage{color}
\usepackage[colorlinks,linkcolor=red]{hyperref}
\usepackage{multirow}
\usepackage{adjustbox}
\usepackage[misc]{ifsym}

\begin{document}
\title{Knowledge Distillation in Federated \\ Edge Learning: A Survey \thanks{Supported by the National Key Research and Development Program of China (No. 2021YFB2900102) and the National Natural Science Foundation of China (No. 62072436, No. 62002346 and No. 61872028).}}
\titlerunning{Knowledge Distillation in Federated Edge Learning}

\author{Zhiyuan Wu\inst{1,2} \and
Sheng Sun\inst{1} \and
Yuwei Wang\inst{1} \textsuperscript{(\Letter)} \and
Min Liu\inst{1,3} \and \\
Xuefeng Jiang\inst{1,2} \and Runhan Li\inst{1,2} \and Bo Gao\inst{4}
}
\authorrunning{Z. Wu et al.}
\institute{Institute of Computing Technology, Chinese Academy of Sciences, Beijing, China \and
University of Chinese Academy of Sciences, Beijing, China \and
Zhongguancun Laboratory, Beijing, China
\and
Beijing Jiaotong University, Beijing, China
\\
\email{\{wuzhiyuan22s,sunsheng,ywwang,liumin, jiangxuefeng21b,lirunhan22s\}@ict.ac.cn\\
bogao@bjtu.edu.cn}
}

\maketitle              % typeset the header of the contribution
\begin{abstract}
The increasing demand for intelligent services and privacy protection of mobile and Internet of Things (IoT) devices motivates the wide application of Federated Edge Learning (FEL), in which devices collaboratively train on-device Machine Learning (ML) models without sharing their private data. 
Limited by device hardware, diverse user behaviors and network infrastructure, the algorithm design of FEL faces challenges related to resources, personalization and network environments. \textcolor{black}{Fortunately}, Knowledge Distillation (KD) has been leveraged as an important technique to tackle the above challenges in FEL. 
In this paper, we investigate the works that KD applies to FEL, discuss the limitations and open problems of existing KD-based FEL approaches, and provide guidance for their real deployment.

\end{abstract}
%
%
%
%%%%%%%%%%此处插入第一张图片

\section{Introduction}
Edge Computing (EC) is an emerging technology deployed in mobile edge networks, which coordinates computing and memory resources in the edge to support low-latency and high-bandwidth-demanding applications for end-side mobile and Internet of Things (IoT) devices.
With the increasing concern for intelligent services and data privacy protection of devices, Federated Edge Learning (FEL) \cite{tak2020federated} is proposed to jointly train Machine Learning (ML) models with decentralized data at the edge of the network.
Unlike conventional centralized training paradigms, FEL keeps \textcolor{black}{private} data on devices and only transmits model parameters or \textcolor{black}{encrypted data information} to the edge server, which is a promising solution for privacy-preserving edge intelligence.
However, FEL faces severe challenges related to resources, personalization, network environments, etc \cite{tak2020federated,yu2021toward,lim2020federated}. 
Shortage of individual device resources and imbalance of resources among devices significantly increase the difficulty of utilizing end-side resources. 
Besides, personalized needs of users call for differentiated on-device models, while the uniform models trained by conventional parameter-averaging-based FEL methods cannot generalize to all devices.
Moreover, non-ideal communication channels and network topology constrain the system design of FEL.

As an ML technique that both enables knowledge transfer and model collaborative training, \textcolor{black}{Knowledge Distillation (KD) \cite{hinton2015distilling,wang2021knowledge} transfers knowledge from one ML model to another, allowing interactive learning among heterogeneous ML models to achieve constructive optimization.} 
Due to flexibility and effectiveness, KD has been applied to solve numerous ML problems, such as model compression \cite{hinton2015distilling,he2019knowledge}, domain adaptation \cite{wu2021spirit,nguyen2021unsupervised}, distributed learning \cite{anil2018large,bistritz2020distributed}, etc.
Recent trends suggest the great potential to address the above challenges in the context of FEL.
Previous works integrating KD into the training process of FEL have been successful in tackling constrained device resources \cite{itahara2021distillation,jeong2018communication,sattler2021cfd,song2022federated,cheng2021fedgems,he2020group,cho2022heterogeneous,wu2023agglomerative}, adapting to heterogeneous devices and user requirements \cite{wu2023fedict,jin2022personalized,zhou2021source,zhang2022fedzkt,yu2022resource,qi2022fedbkd,mishra2021network,wu2024fedcache}, and adapting to complex communication channels as well as network topologies \cite{ahn2019wireless,ahn2020cooperative,oh2020mix2fld,li2021decentralized,taya2022decentralized}. Therefore, a survey is urgently needed to review how KD applies to FEL.

    To the best of our knowledge, \textbf{this paper is the first work to investigate the application of knowledge distillation in federated edge learning.} 
    Different from existing surveys \cite{tak2020federated,yu2021toward,lim2020federated,mora2022knowledge}, we take the challenges faced by FEL as the main clue, introducing existing FEL approaches based on diverse forms of KD techniques and providing guidance for both future research directions and real deployment.
Specifically, the reminders of this paper are organized as follows. 
Section \ref{pre} provides preliminary knowledge of related research directions, including FEL and KD, and elaborates the reasons for concerning KD in FEL. 
Section \ref{challenge} investigates FEL based on KD in addressing resource-constrained, resource-heterogeneous, personalization, non-ideal channels and decentralization challenges in mobile edge networks. 
Section \ref{discuss} summarizes the limitations of existing methods, raises open problems in KD-based FEL research, and provides guidance for real deployment. 
Section \ref{conclude} summarizes the whole paper.

\section{Preliminary}
\label{pre}
\subsection{Federated Edge Learning}
As a practical way to realize edge intelligence, Federated Edge Learning (FEL) implements Federated Learning (FL) systems in mobile edge networks, where massively distributed mobile and Internet of Things (IoT) devices jointly train machine learning models without sharing private data on devices \cite{tak2020federated}. 
Due to diverse user behaviors, limited device capabilities and non-ideal communication environments, FEL faces more severe challenges than conventional FL: more restricted on-device computation and communication resources, more prominent cross-devices data and resource heterogeneity, more complex channel environments, etc \cite{yu2021toward,lim2020federated}. 
How to address the above challenges to accommodate the system infrastructure and user characteristics of mobile edge networks is crucial for practical applications of FEL.

\begin{figure}[!t]
	\centering
	\includegraphics[width=0.85 \textwidth]{
		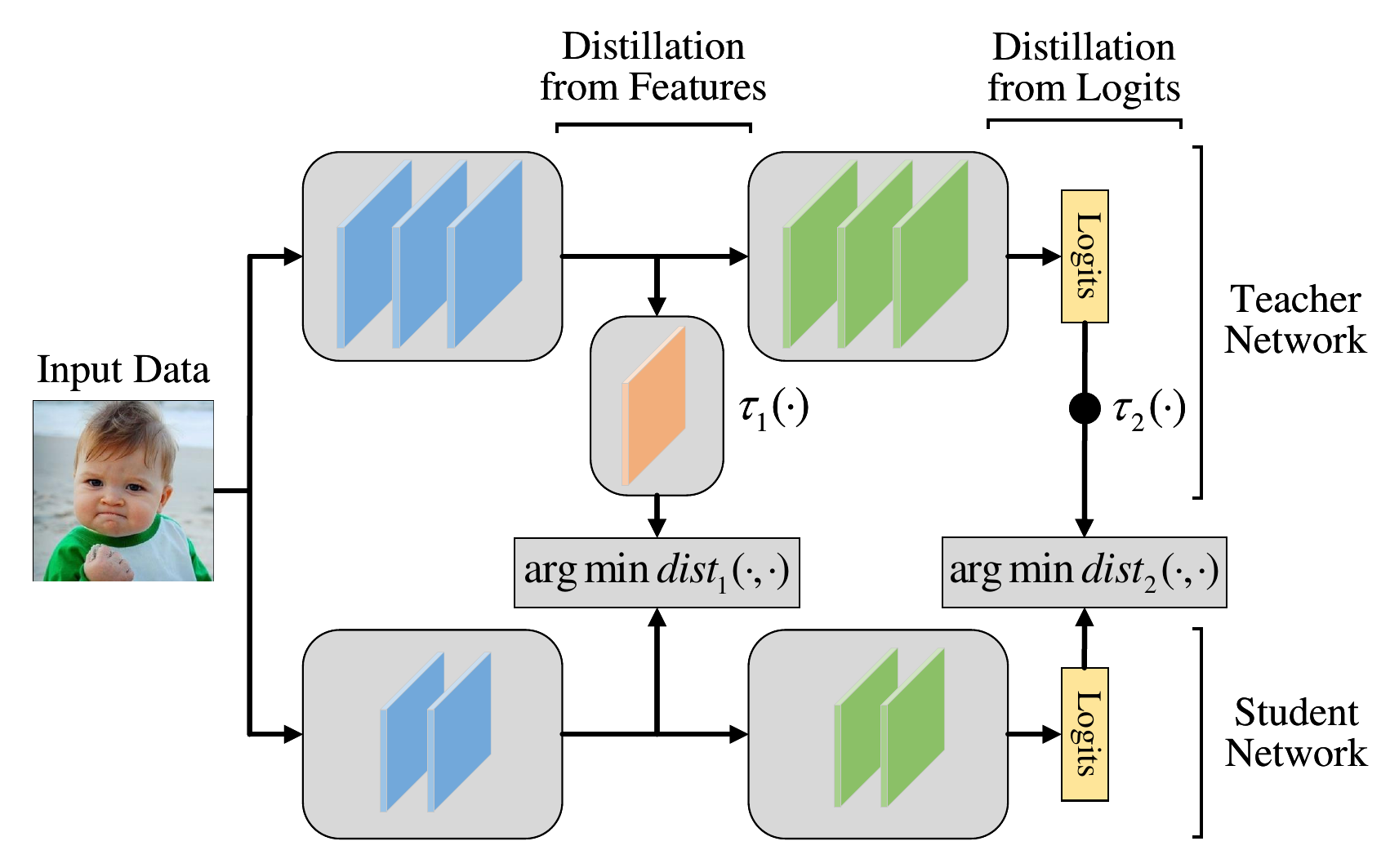}
	\caption{General framework of knowledge distillation.}
    \label{kd}
\end{figure}

\subsection{Knowledge Distillation}
Knowledge Distillation (KD) is a machine learning technique for constructive model training through knowledge transfer \cite{wang2021knowledge,gou2021knowledge}. In common KD frameworks, the transferred model output (typically logits \cite{hinton2015distilling,peng2019correlation} or features \cite{liu2020structured,he2019knowledge,wu2021spirit2}) is referred to as knowledge.
\textcolor{black}{Fig. \ref{kd} shows the general framework of KD.}
The core of KD is that given the same input, one model (student network) simulates the adjusted outputs of another model (teacher network) to learn from the representation of the latter model, that is:
\[\mathop {\arg \min }\limits_{{W_S}} \mathop E\limits_{X\sim{\cal X}} [dist(f(X;{W_S}),\tau (f(X;{W_T})))],\]
where $W_S$ and $W_T$ are model parameters of \textcolor{black}{the student and the teacher networks}, respectively. $f(\cdot;W_*)$ is a nonlinear function determined by parameters $W_*$, which is commonly a neural network. $\tau(\cdot)$ is a knowledge distribution adjustment function for knowledge interpretation, making the transferred knowledge easier to be learned by the student network. 
\textcolor{black}{
During the distillation process, the student network inputs data or features $X\sim \mathcal{X}$, where $\mathcal{X}$ is the set of input data or extracted features. The student network then accepts the knowledge (outputs) extracted by the teacher network on the same input $X$ and achieves knowledge transfer by minimizing the distance $dist(\cdot,\cdot)$ between the two networks' outputs.
}

KD can be customized to solve various machine learning problems, such as model compression via knowledge transfer from bulky models to compact models \cite{wu2021spirit,hinton2015distilling}, distributed model training via knowledge exchange between models \cite{anil2018large,bistritz2020distributed}, etc. 
With elaborate system design, recent studies confirm that KD can be leveraged to solve problems in FL as well \cite{chen2022metafed,li2019fedmd,cho2021personalized,zhang2021parameterized,wu2022exploring,zhu2021data,jiang2022towards,lee2022preservation}.

\subsection{Why Concern Knowledge Distillation in Federated Edge Learning}
KD has achieved success in FL, such as realizing efficient communication \cite{itahara2021distillation,sattler2021cfd}, enabling model heterogeneity \cite{li2019fedmd,zhu2021data} and personalization \cite{zhang2021parameterized,cho2021personalized} among devices, help countering incremental data \cite{dong2022federated,wu2024federated}, etc.
\textcolor{black}{Recent trends suggest that KD has great potential to apply to various learning processes in FEL as an important tool for knowledge transfer or model collaborative training in diversity-constrained mobile edge networks.}
Specifically, the technical characteristics of KD meet the core demands of FEL, and the roles it can play include but not limited to compressing large-scale edge models for on-device deployment \cite{mishra2021network}, transferring local adaptive knowledge to on-device models for personalization \cite{wu2023fedict,jin2022personalized,zhou2021source,wu2024fedcache}, and helping establish novel FL frameworks for enabling heterogeneous device supports \cite{qi2022fedbkd,zhang2022fedzkt,yu2022resource}. 
Representative works that apply KD in FEL are summarized in Table \ref{methods}, \ref{knowledge-and-data}. 
\color{black}
As shown in Table \ref{methods}, prior works are classified into four types according to the role of KD in the FEL: 1) Knowledge transfer, where KD is utilized to transfer knowledge from one ML model to another, occurring only at the end-side or edge-side.
2) Model representation exchange protocol, where on-device and edge models achieve collaborative optimization based on exchanged logits or features. 3) Component of backbone algorithm, where KD-based FEL algorithms are improved as the base algorithm. 
4) Dataset distillation, where small datasets derived from on-device data are synthesized via KD.
In addition, the deployment modes of KD in FEL are classified into edge, end and edge-end collaboration, depending on the application location of KD and whether end-side and edge-side collaboration are required. 
\textcolor{black}{As shown in Table \ref{knowledge-and-data}, some FEL approaches require public datasets, in which edge models and on-device models leverage the same shared public data during distillation to synchronize the knowledge they generate.}
\color{black}

In the following sections, we will detailly introduce the role played by KD in \textcolor{black}{tackling a variety of challenges} faced by FEL.

\begin{table}[!t]
	\caption{Comparison of representative works that apply KD in FEL \textcolor{black}{regarding challenges to tackle, roles play, and deployment modes.}}
	\centering
	\setlength{\tabcolsep}{5pt}
	\renewcommand\arraystretch{1.05}
	\begin{adjustbox}{center}
	\begin{tabular}{l|c|c|c}
		\hline
		\multicolumn{1}{c|}{\textbf{Method}} & \textbf{Challenge to Tackle}                                                     & \textbf{Role of KD in FEL}                                                                             & \textbf{Deployment Mode}                                                          \\ \hline
		Fed-ET \cite{cho2022heterogeneous}                              & \begin{tabular}[c]{@{}c@{}}Limited\\ Computation\end{tabular}                    & \multirow{11}{*}{Knowledge Transfer}                                                                    & \multirow{4}{*}{Edge}                                                             \\ \cline{1-2}
		FedZKT \cite{zhang2022fedzkt}                              & \begin{tabular}[c]{@{}c@{}}Heterogeneous\\ Computation\end{tabular}              &                                                                                                        &                                                                                   \\ \cline{1-2}
		Mix2FLD \cite{oh2020mix2fld}                             & Non-ideal Channel                     &                                                                                                        &                                                                                   \\ \cline{1-2} \cline{4-4} 
		pFedSD \cite{jin2022personalized}                              & \multirow{2}{*}{Personalization}                                                 &                                                                                                        & \multirow{5}{*}{End}                                                              \\ \cline{1-1}
		STU-KD \cite{zhou2021source}                              &                                                                                  &                                                                                                        &                                                                                   \\ \cline{1-2}
		NRFL \cite{mishra2021network}                                & \begin{tabular}[c]{@{}c@{}}Heterogeneous\\ Communication\end{tabular}            &                                                                                                        &                                                                                   \\ \cline{1-2}
		Def-KT \cite{li2021decentralized}                              & Decentralization                                                                 &                                                                                                        &                                                                                   \\ \cline{1-2} \cline{4-4} 
		FedKEM \cite{yu2022resource}                             & \begin{tabular}[c]{@{}c@{}}Heterogeneous\\ Computation\end{tabular}              &                                                                                                        & Edge+End                                                                          \\ \hline
		DS-FL \cite{itahara2021distillation}                               & \multirow{2}{*}{\begin{tabular}[c]{@{}c@{}}Limited\\ Communication\end{tabular}} & \multirow{12}{*}{\begin{tabular}[c]{@{}c@{}}Model Representation\\ Exchange Protocol\end{tabular}}      & \multirow{12}{*}{\begin{tabular}[c]{@{}c@{}}Edge-end\\ Collaboration\end{tabular}} \\ \cline{1-1}
		FD \cite{jeong2018communication}                                  &                                                                                  &                                                                                                        &                                                                                   \\ \cline{1-2}
		FedGEMS \cite{cheng2021fedgems}                              & \multirow{2}{*}{\begin{tabular}[c]{@{}c@{}}Limited\\ Computation\end{tabular}}   &                                                                                                        &                                                                                   \\ \cline{1-1}
		FedGKT \cite{he2020group}                              &                                                                                  &                                                                                                        &                                                                                   \\ \cline{1-2}
		FedBKD \cite{qi2022fedbkd}                              & \begin{tabular}[c]{@{}c@{}}Heterogeneous\\ Computation\end{tabular}              &                                                                                                        &                                                                                   \\ \cline{1-2}
		CMFD \cite{taya2022decentralized}
                              & Decentralization                                                                 &                                                                                                        &                                                                                   \\ \cline{1-2}
        FedCache \cite{wu2024fedcache}                              & \begin{tabular}[c]{@{}c@{}}Personalization+\\Heterogeneous\\ Computation\end{tabular}                                                                  &   &       \\ \cline{1-2}
        FedAgg \cite{wu2023agglomerative}                              & \begin{tabular}[c]{@{}c@{}}Heterogeneous\\ Computation\end{tabular}                                                                  &   &       \\
                              \hline
		FedICT \cite{wu2023fedict}                              & \begin{tabular}[c]{@{}c@{}}Personalization+\\Heterogeneous\\ Computation\end{tabular}                                                                  & \begin{tabular}[c]{@{}c@{}}Knowledge Transfer+\\ Model Representation\\ Exchange Protocol\end{tabular} & \begin{tabular}[c]{@{}c@{}}Edge+End+\\ Edge-end\\ Collaboration\end{tabular}      \\ \hline
		CFD \cite{sattler2021cfd}                                 & \begin{tabular}[c]{@{}c@{}}Limited \\ Communication\end{tabular}                 & \multirow{2}{*}{\begin{tabular}[c]{@{}c@{}}Component of \\ Backbone Algorithm\end{tabular}}          & \multirow{2}{*}{\begin{tabular}[c]{@{}c@{}}Edge-end\\ Collaboration\end{tabular}} \\ \cline{1-2}
		Ahn, et al. \cite{ahn2019wireless,ahn2020cooperative}                         & Non-ideal Channel                         &                                                                                                        &                                                                                   \\ \hline
		FedD3 \cite{song2022federated}                                & \begin{tabular}[c]{@{}c@{}}Limited\\ Communication\end{tabular}                  & Dataset Distillation                                                                                   & End                                                                               \\ \hline
	\end{tabular}
\end{adjustbox}
\label{methods}
\end{table}

\begin{table}[t!]
\caption{\textcolor{black}{Comparison of representative works that apply KD in FEL regarding knowledge types and requirements for public datasets.}}
\centering
\setlength{\tabcolsep}{10pt}
\renewcommand\arraystretch{1.05}
\begin{tabular}{l|c|c}
\hline
\multicolumn{1}{c|}{\textbf{Method}} & \textbf{Knowledge Type}         & \textbf{Require Public Dataset} \\ \hline
STU-KD \cite{zhou2021source}                              & \multirow{16}{*}{Logits}        & \multirow{12}{*}{Yes}        \\ \cline{1-1}
DS-FL \cite{itahara2021distillation}                                &                                 &                              \\ \cline{1-1}
FD \cite{jeong2018communication}                                  &                                 &                              \\ \cline{1-1}
CFD \cite{sattler2021cfd}                                 &                                 &                              \\ \cline{1-1}
FedGEMS \cite{cheng2021fedgems}                             &                                 &                              \\ \cline{1-1}
Fed-ET \cite{cho2022heterogeneous}                              &                                 &                              \\ \cline{1-1}
FedKEM \cite{yu2022resource}                             &                                 &                              \\ \cline{1-1}
FedBKD \cite{qi2022fedbkd}                              &                                 &                              \\ \cline{1-1}
Mix2FLD \cite{oh2020mix2fld}                             &                                 &                              \\ \cline{1-1}
Ahn, et al. \cite{ahn2019wireless,ahn2020cooperative}                         &                                 &                              \\ \cline{1-1}
Def-KT \cite{li2021decentralized}                              &                                 &                              \\ \cline{1-1}
CMFD \cite{taya2022decentralized}                                &                                 &                              \\ \cline{1-1} \cline{3-3} 
pFedSD \cite{jin2022personalized}                              &                                 & \multirow{4}{*}{No}          \\ \cline{1-1}
FedCache \cite{wu2024fedcache}                              &                                 &           \\ \cline{1-1}
FedZKT \cite{zhang2022fedzkt}                              &                                 &                              \\ \cline{1-1}
NRFL \cite{mishra2021network}                                &                                 &                              \\ \hline
FedGKT \cite{he2020group}                              & \multirow{3}{*}{Features+Logits} & \multirow{3}{*}{No}          \\ \cline{1-1}
FedAgg \cite{wu2023agglomerative}                              &  &          \\ \cline{1-1}
FedICT \cite{wu2023fedict}                              &                                 &                              \\ \hline
FedD3 \cite{song2022federated}                               & Distilled Datasets              & No                           \\ \hline
\end{tabular}
\label{knowledge-and-data}
\end{table}

\section{Challenges in Federated Edge Learning with Solutions based on Knowledge Distillation}
\label{challenge}
\subsection{Limited Resources}
%In most FEL settings, the edge server coordinates large-scale mobile devices for collaborative model training. 
As hardware configurations and running power of devices are commonly inferior in FEL, communication and computation resources at the network termination are strictly limited.
Therefore, how to fully utilize available resources to achieve satisfactory system performance is the core issue that KD need to tackle in FEL.

To reduce the communication overhead in FEL, Itahara \cite{itahara2021distillation} proposes DS-FL, which is a distillation-based semi-supervised FL framework. DS-FL achieves efficient communication via on-device local models' outputs exchange between heterogeneous devices \textcolor{black}{and optimizes} local models based on KD with common inputs of an open dataset, thus avoiding model parameters exchange conducted by conventional FEL approaches \cite{jiang2022fedsyl,mills2021multi}.
Jeong \cite{jeong2018communication} employs label-level knowledge alignment in FEL, where the per-label mean logit vectors of each device (knowledge) are periodically uploaded to the edge server. The server then broadcasts the aggregated knowledge to all devices in turn, and the on-device models conduct KD based on the downloaded knowledge for local models' regularization.
Sattler \cite{sattler2021cfd} extends existing KD-based FEL methods \cite{jeong2018communication,sattler2021cfd} through imposing quantization and delta coding to exchanged knowledge, showing $\times 100$ communication efficiency improvement compared with prior works. On top of this, Song \cite{song2022federated} distils local datasets on devices and uploads the distilled dataset to the edge server, requiring only one-shot communication during the entire training process.

To reduce on-device computation overhead in FEL, a series of approaches \cite{he2020group,cheng2021fedgems,cho2022heterogeneous,wu2023agglomerative} are proposed to enable devices training much smaller models than the edge server, leveraging KD as an exchange protocol across model representations. 
Specifically, He \cite{he2020group} and Cheng \cite{cheng2021fedgems} establish alternating minimization FEL frameworks to transfer knowledge from compact on-device models to the large edge model via KD, after which the on-device models are optimized based on the knowledge transferred back from the edge. Wu \cite{wu2023agglomerative} proposes a federated large-model training framework based on mutual distillation over generated bridge samples. In addition, Cho \cite{cho2022heterogeneous} proposes an ensemble knowledge transfer algorithm for collaborative training across devices, where the large edge model learns the weighted consensus of the uploaded small on-device models via KD.

\subsection{Heterogeneous Resources}
\textcolor{black}{Resources among devices exist huge divergence in FEL} deriving from disparate hardware configurations, cruising ability, and network connectivity of different devices. 
Resource gap among devices poses requirements that KD-based FEL approaches should support heterogeneous training demands depending on the hardware and power of devices, and tolerate disparate communication capabilities of devices.

Existing works employ heterogeneous on-device models to adapt to the computing power of heterogeneous devices' hardware, and leverage KD to transfer knowledge from the shared edge model to heterogeneous on-device models.
Specifically, Zhang \cite{zhang2022fedzkt} proposes a zero-shot knowledge transfer approach that adversarially trains a generator with uploaded on-device models. After that, pseudo-data generated by the trained generator is applied to a bi-directional KD process across the edge and the devices, aiming to integrate the knowledge from heterogeneous on-device models to the edge server and guide the on-device models to achieve more generalized training performance in turn. Nguyen \cite{yu2022resource} adopts a variant of co-distillation \cite{anil2018large} at the network termination for collaboratively optimizing on-device models and the downloaded edge model, while integrating the knowledge from heterogeneous devices via ensemble KD in the edge. 
In addition, Qi \cite{qi2022fedbkd} employs KD after on-device model uploading and edge model downloading, and uses a conditional variational autoencoder to complement both the required public data and private data.
It is also worth noting that previously mentioned approaches taking KD as the end-edge model representation exchange protocol \cite{he2020group,wu2024fedcache,cheng2021fedgems,cho2022heterogeneous,wu2023fedict,itahara2021distillation,jeong2018communication,wu2023agglomerative} also support heterogeneous on-device models, which are partly applicable on heterogeneous devices as well.

Another line of research addresses the problem of unequal available bandwidth among devices in FEL, where Mishra \cite{mishra2021network} proposes a network resource-aware FL approach using KD. Precisely, devices are grouped depending on their available bandwidth resources and are set models that accommodate their bandwidth. During the training process, devices accept the initialized on-device model parameters from the edge server and transfer back the locally trained models in groups from large to small. 
The edge server iteratively distributes the aggregated on-device models to their corresponding groups of devices and uses KD to compress the uploaded models for initializing model parameters of the next group of devices.

\subsection{Personalization}
Local data of different devices are generated from the daily use of their respective users. 
Due to differentiated user behaviours, data among devices tend to exhibit non-independent identically distributed (non-IID), and will make it difficult for the trained edge-side global model to generalize to all devices. 
Therefore, techniques for global model personalization are urgently needed to address the challenges posed by non-IID data in FEL.

In \cite{wu2023fedict}, a novel bi-directional knowledge distillation framework practical for multi-access edge computing is proposed to achieve personalized optimization of on-device models while ensuring fast convergence of the global model. In \cite{wu2024fedcache}, a knowledge cache-driven federated learning architecture is proposed that guarantees satisfactory personalized performance while conforming to devices-side limitations in edge computing. Besides, Jin \cite{jin2022personalized} proposes to reserve personalized models on devices. At the end of each communication round, self-distillation is employed on devices to transfer knowledge from the personalized on-device models of the previous round, and the distilled model is retained to the next round for further personalized optimization. In addition, Zhou \cite{zhou2021source} models FEL as a federated domain adaptation problem, and leverages distillation-based source-free unsupervised domain adaptation to transfer the knowledge from the edge server in a memory-efficient manner, aiming to achieve high inference accuracy in local data environments on edge devices with acceptable training cost.

\subsection{Non-ideal Channels}
Most existing KD-based FEL approaches assume that communication links between devices and the edge server are ideal channels, which are impractical in mobile edge networks in reality. 
For instance, the capacity of uplink and downlink channels are often unequal in real network environments, or the FEL system usually is deployed on fading channels, etc.
Such wireless communication constraints present new challenges to implementing KD-based FEL.

In \cite{oh2020mix2fld}, Oh addresses uplink-downlink capacity asymmetry by allowing devices to upload model outputs but download model parameters during training, and the edge model conducts optimization via KD, with uploaded mixed samples from devices as input.
Ahn \cite{ahn2019wireless} models devices' uplink as gaussian multiple-access channels, and provides digital implementations with separate source-channel coding and over-the-air computing implementations with joint source-channel coding for KD-based FEL. Further, the case in which uplink is a multiple access fading channel and downlink is a fading broadcast channel is considered in \cite{ahn2020cooperative}, and concrete implementations for KD-based FEL are given as well.

\subsection{Decentralization}
In mobile edge networks, locating powerful and reliable edge servers can be challenging due to weak infrastructure, restricted network topologies, and a lack of trust in remote servers.
Therefore, decentralization of FEL is a promising research direction in which KD provides a tool for stable convergence of decentralized FEL systems with heterogeneous device support.

Taya \cite{taya2022decentralized} proposes a decentralized FL algorithm for Internet of Everything (IoE) in multi-hop networks, where each device learns the transferred knowledge from neighboring devices via KD. In addition, Li \cite{li2021decentralized} leverages mutual knowledge transfer to allow clients to learn knowledge from each other, enabling training on heterogeneous datasets with client-drift prevented.

\section{Discussion}
\label{discuss}
\subsection{Limitations}
Although KD-based FEL has witnessed a phased success, quite a few approaches \cite{he2020group,itahara2021distillation,jeong2018communication,sattler2021cfd,oh2020mix2fld,ahn2019wireless} only achieve comparable or even worse performances than inchoate FL methods represented by FedAvg \cite{mcmahan2017communication}. 
On the one hand, the amount of useful information transferred from knowledge in a single communication round is less than model parameters, and the gain in communication efficiency achieved by the above methods is at the cost of inferior on-device model performance.
On the other hand, adopting KD as model representation exchange protocols implicitly tolerates greater divergence of on-device models during training, which results in models receiving knowledge irrelevant to their own training goals \cite{cho2021personalized,zhang2021parameterized}.
Therefore, it is necessary to reconstruct the transferred knowledge so that on-device models can capture more useful information and better exploit the unique advantages of KD-based FEL with acceptable performance in mobile edge networks.

\subsection{Open Problems}
Since the related research is still at an early stage, there are many open problems in KD-based FEL. 
\textcolor{black}{In terms of device connectivity,} it is impractical to keep devices online all the time, and how to cope with devices offline and dropout in KD-based FEL systems remains unsolved. 
\textcolor{black}{Considering training acceleration,} how to schedule devices in KD-based FEL to achieve efficient asynchronous training has so far never been studied.
\textcolor{black}{When it comes to incentive mechanism, ways of} measuring device contributions in the KD-based FEL and creating incentives to keep devices consistently motivated to participate in FEL training is also worth attention.
\textcolor{black}{Discussing about privacy protection,} how to encode knowledge in KD-based FEL systems for counteracting inversion attacks is also worth concerns, especially for methods that upload features \cite{he2020group,wu2023fedict} or transformed data \cite{oh2020mix2fld,song2022federated} during training which are relatively easy to be attacked.
\textcolor{black}{Last but not least,} many prior works \cite{zhou2021source,itahara2021distillation,sattler2021cfd,cheng2021fedgems,cho2022heterogeneous,yu2022resource,taya2022decentralized} rely on a publicly available dataset that is inaccessible in real scenarios \cite{yu2021toward,wu2023fedict}, \textcolor{black}{thus novel KD-based FEL methods should provide normative solutions.}

\subsection{Practical Guidance for Real Deployment}
Applying KD-based FEL in practice is a multi-dimensional problem that requires ensembled techniques with performance trade-offs. We suggest that a practical FEL system with KD could include but not limited to resource-aware FEL architecture, model-agnostic representation exchange protocols without open datasets, knowledge adaptation and refinement, KD-specific knowledge compression, privacy-preserving knowledge transfer, techniques for learning from knowledge, solutions for complex communication channels in mobile edge networks, etc.
Only in this way can FEL be improved by KD, with wider applications in real-world scenarios.

\section{Conclusion}
\label{conclude}
Taking the challenges faced by Federated Edge Learning (FEL) as the main clue, this paper surveys prior works on applying Knowledge Distillation (KD) to FEL, and classifies the role of KD in FEL into four types of knowledge transfer, model representation protocol, component of backbone algorithm and dataset distillation. \textcolor{black}{In addition, the deployment modes of KD are categorized as edge-side, end-side, and edge-end collaboration.}
We also provide guidance for future research directions and real deployment of KD-based FEL.

In principle, KD can effectively help address the core challenges in FEL and be deemed a functional tool for knowledge transfer and model collaborative training. 
In addition, we suggest that future research on KD-based FEL should focus on improving model accuracy, while paying attention to open problems we raised. Real deployment requires reasonable integration of existing techniques and multifaceted performance trade-offs.

~\\
\noindent
\textbf{Acknowledgment.} We thank Hui Jiang, Yuchen Zhu and Tianliu He from Institute of Computing Technology, Chinese Academy of Sciences for inspiring suggestions.

 \bibliographystyle{unsrt}

\end{document}